\documentclass[runningheads]{llncs}
\usepackage{graphicx,amsmath,booktabs,pifont,multirow}
\begin{document}
\title{Capturing Temporal Components for Time Series Classification}
%
%
\author{Venkata Ragavendra Vavilthota\inst{1} \and
Ranjith Ramanathan\inst{2} \and
Sathyanarayanan N. Aakur\inst{3}\orcidID{0000-0003-1062-8929}}
\authorrunning{Vavilthota et al.}
%
\institute{Department of Computer Science, Oklahoma State University, Stillwater OK 74078, USA \and
Department of Animal and Food Sciences, Oklahoma State University, Stillwater OK 74078, USA \and
Department of Computer Science and Software Engineering, Auburn University, Auburn, 36849\\
\email{san0028@auburn.edu}}
\maketitle              
\begin{abstract}
Analyzing sequential data is crucial in many domains, particularly due to the abundance of data collected from the Internet of Things paradigm. Time series classification, the task of categorizing sequential data, has gained prominence, with machine learning approaches demonstrating remarkable performance on public benchmark datasets. However, progress has primarily been in designing architectures for learning representations from raw data at fixed (or ideal) time scales, which can fail to generalize to longer sequences. This work introduces a \textit{compositional representation learning} approach trained on statistically coherent components extracted from sequential data. 
Based on a multi-scale change space, an unsupervised approach is proposed to segment the sequential data into chunks with similar statistical properties. 
A sequence-based encoder model is trained in a multi-task setting to learn compositional representations from these temporal components for time series classification. 
We demonstrate its effectiveness through extensive experiments on publicly available time series classification benchmarks. Evaluating the coherence of segmented components shows its competitive performance on the unsupervised segmentation task.
\end{abstract}
\section{Introduction}\label{sec:intro}
Time series data is ubiquitous in many domains, such as healthcare~\cite{ramnath2020smart} and robotics~\cite{trehan2022towards}. 
Given the widespread presence of sensors and smart devices, abundant sequential (time series) data across different domains has been collected, giving rise to several important tasks in time series analysis, such as classification, segmentation, and anomaly detection. 
Time series classification is one task that has received significant attention in recent years. The goal is to learn robust features from sequential data to classify them into their respective categories. 
Machine learning approaches~\cite{schafer2015boss,lucas2019proximity}, particularly deep learning approaches~\cite{tang2021omni,yue2022ts2vec}, have shown tremendous progress in learning models for time series data classification and have resulted in interesting applications such as sleep state segmentation~\cite{ramnath2020smart} and pandemic modeling~\cite{dash2021intelligent}, to name a view. 

The sequential nature of the time series data offers several challenges for classification. First, the sequence length can vary across samples within categories, which requires learning representations robust to such intra-class variations. Second, understanding the ideal time scale for extracting meaningful patterns is challenging, primarily caused by measurement errors and phase/amplitude changes across samples. Finally, long-duration sequences can have dependencies that span different time scales and pose a significant challenge to representation learning approaches. While driving tremendous progress, learning from raw signals relies heavily on representation learning mechanisms to capture intricate, compositional properties for tackling these challenges. Explicitly capturing the underlying temporal structure of signals in the representation can help alleviate this dependency and lead to more robust performance on downstream tasks. Such representations have shown tremendous potential in scene recognition tasks~\cite{locatello2020object} by considering objects as atomic components that combine to compose the overall scene. However, time series may not have such clear distinctions for recognizing boundaries between components, requiring a novel paradigm for defining and detecting temporal components in sequential data.

\begin{figure}
    \centering
    \includegraphics[width=0.65\columnwidth]{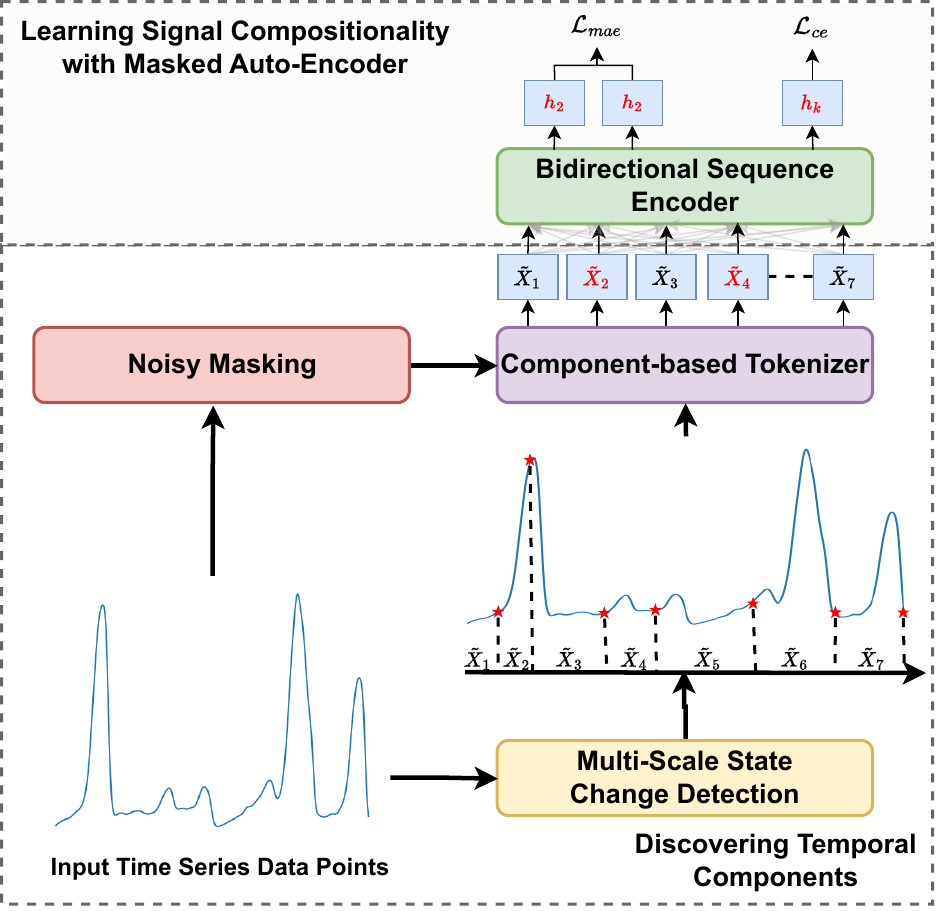}
    \caption{\textbf{Overall architecture} of the proposed approach is illustrated here. First, we introduce a multi-scale state change detection model to segment sequential data into components and then use a sequence-based encoder to learn compositional representations for time series classification.}
    \label{fig:Arch}
\end{figure}

In this work, we propose to capture the different atomic components that combine to form these signals in a compositional representation. We consider a time series data point, or signal, to be a sequence of data points ordered by some condition and can be segmented into chunks that share semantic or statistical properties. 
These chunks, or sub-series, are called \textit{components} of the overall signal. 
Rather than learning representations over the raw sequential data, representations from this sequence of \textit{components} can result in a compositional feature that can span longer durations with reduced computational complexity. The overall approach is illustrated in Figure~\ref{fig:Arch}. We first establish a multi-scale change space (Section~\ref{sec:BIC}) to segment (or tokenize) the signal into components at different temporal scales. Then, we learn compositional representations (Section~\ref{sec:BiLSTM}) from these segments in a multi-task learning setting. Extensive evaluation (Section~\ref{sec:results}) on publicly available benchmark datasets shows that the approach performs competitively with state-of-the-art approaches and scales well to longer duration time series data. These components are remarkably similar to natural segments found in time series data (Section~\ref{sec:segmentation}), and the approach can naturally be extended to unsupervised time series segmentation. Without bells and whistles, the approach performs competitively to state-of-the-art techniques designed explicitly for segmentation and outperforms other non-learning-based methods. 

The \textbf{contributions} of our approach are four-fold: (i) we are, to the best of our knowledge, to introduce a multi-scale change space for time series data to segment them into statistically atomic components, (ii) we introduce the notion of compositional feature learning from temporally segmented components in time series data rather than modeling the raw data points, (iii) we show that the temporal components detected by the algorithm are highly correlated with natural boundaries in time series data by evaluating it on the time series segmentation task, achieving state-of-the-art performance compared with other non-learning-based approaches, and (iv) we establish a competitive baseline that provides competitive performance with the state-of-the-art approaches on benchmark datasets for both time series classification and segmentation with limited training needs and without explicit handcrafting. 

We structure the paper as follows. We review the relevant literature and techniques used in this work in Section~\ref{sec:related_work}, followed by an overview and detailed explanation of the proposed approach in Section~\ref{sec:proposed}. We present and analyze the quantitative results in Section~\ref{sec:results} and demonstrate how it can be expanded to tackle other time series analysis tasks such as unsupervised segmentation in Section~\ref{sec:segmentation}. Finally, in Section~\ref{sec:conclusion}, we discuss its limitations and future directions. 
\section{Related Work}\label{sec:related_work}
\textbf{Time series classification} has been tackled through three major types of approaches. 
\textit{Classical} approaches, such as those based on handcrafted feature learning~\cite{schafer2015boss,lucas2019proximity,lines2018time,hills2014classification}, have attempted to learn discriminative features from modeling the time series at different scales through techniques such as shapelet transforms~\cite{lines2018time,hills2014classification}, distance-based transforms~\cite{lucas2019proximity,chen2013dtw}, and bag-of-symbols~\cite{schafer2015boss,shifaz2020ts}, to name a few. However, their computational complexity increases almost exponentially as the duration of the time series increases, and hence, they appear to hit a wall of scalability. 
\textit{Deep learning}-based approaches, using architectures such as convolutional neural networks (CNNs)~\cite{kiranyaz20211d} and transformers~\cite{vaswani2017attention}, have opened a wave of large models pre-trained on significant amounts of data~\cite{kenton2019bert}. Deep learning models have focused on modeling the data at the ideal time scale~\cite{tang2021omni,cui2016multi} for capturing robust representations using different backbones such as CNNs~\cite{yue2022ts2vec,zheng2016exploiting,zhao2017convolutional,serra2018towards,franceschi2019unsupervised,wang2017time}, recurrent neural networks (RNNs)~\cite{tanisaro2016time,tonekaboni2020unsupervised}, and transformers~\cite{zerveas2021transformer,eldele2021time}. 
\textit{Ensemble-based} approaches~\cite{tang2021omni,lines2018time,shifaz2020ts,dempster2020rocket}, i.e., using multiple predictions from the different aspects of the same time series data, have made significant strides in establishing the state-of-the-art performance on several benchmark datasets~\cite{bagnall2017great,dau2019ucr,bagnall2018uea}. Our work, however, offers a novel framework to capture multi-scale representations by detecting temporal components at different time scales and integrating them in a unified representation without the need for ensembling and additional overhead in the form of annotations. 

Approaches to \textbf{time series segmentation} have primarily focused on detecting boundaries in sequential data through heuristic-based, domain-specific approaches. Broadly categorized into three categories~\cite{truong2020selective}, the time series is segmented by comparing the features of consecutive fixed-size windows using their likelihood of belonging to the same segment~\cite{Kawahara2021}, assessing homogeneity using kernels~\cite{harchaoui2009regularized} or mapping them into graph-based representation for extracting sub-graphs (segments) through heuristics such as pairwise similarity~\cite{chen2023graph}. Search-based approaches~\cite{sen1975tests,adams2007bayesian,draayer2021reevaluating} and learning-based approaches~\cite{schafer2021clasp,gharghabi2019domain,deldari2020espresso} have offered a way forward to domain-agnostic segmentation by learning sequence-level representations and segmenting them based on similarity measures. The former assigns costs to plausible boundaries and finds optimal segments by minimizing these costs, while the latter focuses on learning boundaries through pre-text tasks as self-supervision. Many of these approaches require the number of segments to be pre-defined, with learning-based approaches such as ClaSP~\cite{schafer2021clasp} being a notable exception. 
Our approach built on BIC-based tokenization (Section~\ref{sec:BIC}) belongs to the search-based approach category and performs \textit{domain-agnostic} segmentation by comparing statistical similarity measures without training. 

\section{Proposed Framework}\label{sec:proposed}
In this section, we outline the proposed framework to extract temporal components from sequential data and learn a robust, compositional representation in a multi-task setting. We first outline the problem formulation to provide an overview of the approach and then introduce the multi-scale change space used to discover temporal components in signals. Finally, we introduce the representation learning mechanism used to combine these temporal components into a robust representation.  

\textbf{Problem Formulation.} We address the task of classifying univariate time series data by decomposing the signal into its constituent parts. We aim to characterize and build a rich signal representation by detecting parts (sub-series) that compose the overall signal. Inspired by theories of compositional event understanding~\cite{zacks2001event}, we consider these parts atomic, i.e., each sub-series cannot be broken down into smaller components. To this end, we consider a multi-scale approach to identify these components at different time scales to account for the unique challenges inherent in time series data, such as variations that are introduced during data collection~\cite{bagnall2017great,dau2019ucr} (i.e., sampling rate and record length) and unavoidable intra-class variations (such as amplitude offset and warping). Following prior works on state-spaces~\cite{laptev2001multi,krishnan2010detecting}, we define a signal-dependent change scale-space that captures the multi-scale structure of the signal based on its temporal change points. The overall architecture is illustrated in Figure~\ref{fig:Arch}. First, we identify the temporal change points in the signal using a statistics-based multi-scale organization (Section~\ref{sec:BIC}), which allows us to break the signal down into its components. Second, we learn compositional relationships from these signal components using a bidirectional sequence learning model (Section~\ref{sec:BiLSTM}) trained in a multi-task setting. Combined, these two steps help identify atomic components in time series signals and help capture their temporal structure in a purely bottom-up fashion without auxiliary data.

\subsection{Discovering Temporal Components of Signals}\label{sec:BIC}
The first step in our approach is to discover temporal sub-components that compose time series signals. These sub-components are temporal chunks whose statistics (mean, variance, etc.) are consistent within the sub-series yet vary significantly with neighboring chunks. Hence, detecting the change in statistics at multiple time scales allows us to discover these temporal components in univariate signals. 
We use the premise from statistics-based speaker-turn detection approaches~\cite{chen1998speaker,krishnan2010detecting} to define a function TSCS (Time Series Change Space) to capture the temporal change space in time series data ($X_{0,N}=\{x_1, x_2, x_3, \ldots, x_N\}$). It is a two-dimensional function over time ($t$) and temporal scale ($\delta$) that characterizes the varying statistics between two sub-series $t{-}\delta$ and $t{+}\delta$ to detect a possible temporal change point (time series component) at time $t$, given a temporal scale $\delta$. 
We cast this formulation as a hypothesis-testing problem. The null hypothesis is that two consecutive chunks are different and thus require two different models to represent them individually. The alternative hypothesis is that they are very similar and belong to a single, longer chunk one model can represent. We evaluate each hypothesis by fitting a single Gaussian model~\cite{chen1998speaker} for the chunks from each hypothesis. 
Hence, the difference in the Bayesian Information Criterion (BIC) between the two models at time $t$ provides a measure of their separability based on their statistics. 
Formally, we define the state space ($TSCS(t,\delta)$) as a function of BIC given by 
\begin{equation}
\begin{split}
     TSCS(t,\delta) &= \frac{\delta}{2}(log\vert\sigma_{X_{t-\delta},t} \vert + log\vert\sigma_{X_{t,t+\delta}} \vert) \\
     &-\delta (log\vert\Sigma_{X_{t-\delta, t+\delta}}\vert) + \delta P
\end{split}
    \label{eqn:BIC}
\end{equation}
where $log\vert\sigma_{X_{t-\delta},t}\vert$ and $log\vert\sigma_{X_{t,t+\delta}}\vert$  refer to the BIC of the single Gaussian representation for the subseries from time $t-\delta$ to $t$ and from $t$ to $t+\delta$, respectively; $log\vert\Sigma_{X_{t-\delta, t+\delta}}\vert$ refers to the BIC of a multivariate jointly considering both subseries; and $P$ is a penalty term to account for the size of the subseries considered and is typically set to $log(T)$ where T is the length of the subseries considered. Higher values of TSCS indicate that the two sub-series are separate components, i.e., a change in statistics is likely and indicates the presence of a change point.

Given this change space, we can build a multi-scale representation by varying the time scale $\delta$ over a range and summing up the resulting BIC curves. Formally, this can be defined as 
\begin{equation}
MS{-}TSCS(t) = \sum_{\delta \in \Delta}{TSCS(t,\delta)} 
\end{equation}
where $\Delta$ is the set of all time scales for detecting time series components. In practice, we consider $\Delta$ to range from $10$ time steps to $500$ time steps. We then pass the curve from $MS{-}TSCS(t)$ through a low pass filter to extract peaks that provide possible time steps to segment the time series. 
We select peaks with high \textit{saliency}, i.e., if it is more than two standard deviations from its neighbors.
This is a common approach in statistics-based outlier detection literature~\cite{chen1998speaker,krishnan2010detecting} and provides a good measure of temporal saliency for this problem. Given the temporal change locations, the \textit{ideal} number of segments per dataset is computed as the average number of components across classes in the training set. 
We find that considering too few (or smaller) values in $\Delta$ will result in fewer segments and poor representations. Note that not all time series will have such components that are statistically separable. We use a uniform sampling approach to split the series into $15$ equal segments in these cases. Empirically, segmenting chunks into more than $50$ segments is not ideal and could degrade the performance, particularly on smaller datasets. 

\subsection{Capturing Signal Compositionality}\label{sec:BiLSTM}
The second step in our approach is to learn robust representations from the multi-scale components extracted using the MS-TSCS function defined in Section~\ref{sec:BIC}. Given the ideal number of segments $K$, the input sequence is \textit{tokenized} $X_N{=}\{x_1, x_2, x_3, \ldots, x_N\}$ into its constituent segments $\tilde{X}_K{=}\{\tilde{X}_1,\tilde{X}_2,\tilde{X}_3,\ldots,\tilde{X}_k\}$. For capturing \textit{compositional} representations, we then use a masked auto-encoding loss function~\cite{bao2020unilmv2} to train the encoding model (with parameters $\Theta$). The masked auto-encoding loss randomly masks $M<k$ components and forces the encoder to independently predict the masked components by conditioning on the context provided by the unmasked components. Given the tokenized time series data $\tilde{X}_K{=}\{\tilde{X}_1,\tilde{X}_2,\tilde{X}_3,\ldots,\tilde{X}_k\}$ and masked components $M{=}\{m_1,m_2,\ldots,m_{\vert M\vert}\}$, the masked auto-encoding loss is 
\begin{equation}
    \mathcal{L}_{mae} = -\sum_{X_i\in \mathcal{C}} log \prod_{m\in M}p(\tilde{X}_m\vert\tilde{X}_{K\setminus M})
    \label{eqn:loss_MAE}
\end{equation}
where $p(\tilde{X}_m\vert\tilde{X}_{K\setminus M})$ is the probability of predicting the randomly masked components in set $\{M\}$. This probability is computed as the mean squared error over the masked component's values. We use a bidirectional LSTM~\cite{hochreiter1997long} as our encoder and ensure that the mask is bidirectional, i.e., the context for predicting the masked component is present on both sides of the mask. This masking procedure has successfully been used to train text-based~\cite{kenton2019bert} and image-based~\cite{xie2023data} encoders. We extend the formulation to univariate time series data. The hidden states of the forward and backward LSTM cells, $h^f_t$ and $h^b_t$, respectively, are concatenated and used as the feature representation for time series classification, optimized by the cross-entropy loss ($\mathcal{L}_{CE}$). Hence, the overall objective function is given by
\begin{equation}
\mathcal{L}_{tot} = \lambda_1\mathcal{L}_{mae} + \lambda_2\mathcal{L}_{CE}
\label{eqn:overall_loss}
\end{equation}
where $\lambda_1$ and $\lambda_2$ are tunable parameters that trade-off between the two losses. The values of $\lambda_1$ and $\lambda_2$ are varied according to a pre-set schedule to balance the representation learning capabilities from the self-supervised masked auto-encoding loss ($\mathcal{L}_{mae}$) and the discriminative, class-specific properties imbued by the supervised cross-entropy loss ($\mathcal{L}_{ce}$). 

\textbf{Implementation Details.} We use a bidirectional LSTM model with a hidden size of 160 neurons, followed by a dense layer with 320 neurons, as our encoder architecture. The ReLU activation is used for all layers. All segmented components are padded as necessary to be equal in length. We use $5\%$ of the training data for validation. 
We use the same pre-processing as previous work~\cite{yue2022ts2vec}. $\lambda_1$ and $\lambda_2$ are varied as follows: for the first 100 epochs, $\lambda_1=1$ and $\lambda_2=0$, then $\lambda_1=2$ and $\lambda_2=1$. The network is trained for 250 epochs or until convergence, i.e., the loss does not improve on the validation set. All experiments were conducted on a workstation server with a 32-core AMD ThreadRipper CPU, 128 GB RAM, and an NVIDIA RTX 3060. 

\section{Experimental Evaluation}\label{sec:results}
In this section, we present the results from the experimental evaluation of the proposed approach. We begin with a discussion on the experimental setup, followed by the quantitative results, and conclude with a qualitative discussion on the representations learned by the approach. 
\subsection{Experimental Setup}\label{sec:data}
\textbf{Data.} We evaluate the proposed approach on 85 datasets collated in the UCR time series archive~\cite{bagnall2017great}. It consists of univariate time series datasets collected from different sensors and domains such as health care, speech reorganization, and spectrum analysis, to name a few. The archive provides a comprehensive benchmark for evaluating time series classification models~\cite{tang2021omni,yue2022ts2vec,franceschi2019unsupervised} across diverse datasets with varying characteristics. The number of classes in each dataset ranges from 2 to 6, the number of time steps per sample varies from 24 to 2709, and the number of training samples per dataset from 16 to 8926. Additionally, we evaluate the approach on 15 datasets with the longest timesteps from the UCR-85~\cite{bagnall2017great} and the UCR-128~\cite{dau2019ucr} datasets to evaluate its ability to capture robust representations from time series with longer duration. We use the official train and test splits on all datasets for a fair comparison with prior works. Average accuracy across all datasets is used to quantify the performance on the UCR time series archive. 
Code and performance for baselines are obtained from publicly available implementations of prior works~\cite{tang2021omni,yue2022ts2vec}.

\begin{table}[t]
\centering
\caption{\textbf{Performance evaluation} of the proposed approach with  state-of-the-art approaches on 85 datasets from the UCR time series archive~\cite{bagnall2017great,dau2019ucr}}
\begin{tabular}{|l|c|c|c|}
\hline
\textbf{Approach} & \textbf{Ensemble?} & \textbf{Backbone} & \textbf{Accuracy} \\ \hline
TST~\cite{zerveas2021transformer} & \ding{55} & Transformer & 64.901\\\hline
MCDCNN~\cite{zheng2016exploiting} & \ding{55} & CNN & 68.551\\\hline
TWIESN~\cite{tanisaro2016time} & \ding{55} & RNN & 68.636\\\hline
TS-Encoder~\cite{serra2018towards} & \ding{55} & CNN & 71.909\\\hline
Time-CNN~\cite{zhao2017convolutional} & \ding{55} & CNN & 72.284\\\hline
DTW~\cite{chen2013dtw} & \ding{55} & Distance & 74.040\\\hline
TS-TCC~\cite{eldele2021time} & \ding{55} & CNN-Transformer & 77.764\\\hline
TNC~\cite{tonekaboni2020unsupervised} & \ding{55} & Bi-RNN & 77.896\\\hline
PF~\cite{lucas2019proximity} & \ding{55} & Distance & 80.419\\\hline
T-Loss~\cite{franceschi2019unsupervised} & \ding{55} & Dilated CNN & 80.482\\\hline
BOSS~\cite{schafer2015boss} & \ding{55} & Bag of Symbols & 81.019\\\hline
FCN~\cite{wang2017time} & \ding{55} & CNN & 81.634\\\hline
ResNet~\cite{wang2017time} & \ding{55} & CNN & 82.201\\\hline
ST~\cite{hills2014classification} & \ding{55} & Shapelets & 82.236\\\hline
TS2Vec~\cite{yue2022ts2vec} & \ding{55} & Dilated CNN & 82.934\\\hline
Ours & \ding{55} & Bi-RNN & \textbf{83.309}\\\hline\hline
TS-CHIEF~\cite{shifaz2020ts} & \ding{51} & Bag of Symbols & 84.641\\\hline
HIVE-COTE~\cite{lines2018time} & \ding{51} & Multiple & 84.714\\\hline
OS-CNN~\cite{tang2021omni} & \ding{51} & CNN & 84.774\\\hline
ROCKET~\cite{dempster2020rocket} & \ding{51} & CNN & \textbf{85.077}\\\hline

\end{tabular}

\label{tab:sota}

\end{table}

\textbf{Baselines.} We compare against state-of-the-art univariate time series classification models, which use different representation learning backbones and propose robust learning methods to account for high intra-class variation common in time series data. Chiefly, we compare against models with CNN backbones~\cite{yue2022ts2vec,zheng2016exploiting,zhao2017convolutional,serra2018towards,franceschi2019unsupervised,wang2017time}, transformer backbones~\cite{zerveas2021transformer,eldele2021time}, RNN backbones~\cite{tanisaro2016time,tonekaboni2020unsupervised}, and other hand-crafted features such as shapelet transforms~\cite{hills2014classification}, distance-based metrics~\cite{lucas2019proximity,chen2013dtw}, and bag-of-symbols~\cite{schafer2015boss}. We also compare against ensembles~\cite{tang2021omni,lines2018time,shifaz2020ts,dempster2020rocket}, which explicitly capture representations at multiple time scales, which can require additional overhead for training. 

\begin{table}[t]
\centering
\caption{\textbf{Performance on 15 longest sequence} time series data from the UCR Archives~\cite{bagnall2017great}, compared against state-of-the-art models with different backbones. 
}
\resizebox{\columnwidth}{!}{
\begin{tabular}{|l|c|c|c|c|c|c|}
\hline
 \multicolumn{1}{|c|}{\textbf{Backbone$\rightarrow$ }} & \multicolumn{2}{c|}{\textbf{CNN}} & \multicolumn{2}{c|}{\textbf{Transformer}} & \multicolumn{2}{c|}{\textbf{Bi-RNN}} \\ \cline{2-7} 
\multicolumn{1}{|c|}{\textbf{Dataset $\downarrow$}}	&  \textbf{TS2Vec~\cite{yue2022ts2vec}}   &   \textbf{OS-CNN~\cite{tang2021omni}}   &  \textbf{TS-TCC}~\cite{eldele2021time}  &   \textbf{TST}~\cite{zerveas2021transformer}   &  \textbf{TNC}~\cite{tonekaboni2020unsupervised}   &    \textbf{Ours}    \\ \hline
Rock & \textbf{70.00} & 55.00 & 60.00 & 68.00 & 58.00 & \textbf{70.00}\\\hline
HandOutlines & 92.20 & 92.95 & 72.40 & 73.50 & 93.00 & \textbf{94.05}\\\hline
HouseTwenty & 91.60 & \textbf{94.87} & 79.00 & 81.50 & 78.20 & \underline{92.44}\\\hline
InlineSkate & \underline{41.50} & \textbf{42.92} & 34.70 & 28.70 & 37.80 & 41.09\\\hline
EthanolLevel & 46.80 & 73.08 & 48.60 & 26.00 & 42.40 & \textbf{87.00}\\\hline
SemgHandSubjectCh2 & \textbf{95.10} & 71.84 & 75.30 & 48.40 & 77.10 & \underline{91.56}\\\hline
SemgHandGenderCh2 & \textbf{96.30} & 85.61 & 83.70 & 72.50 & 88.20 & \underline{89.33}\\\hline
SemgHandMovementCh2 & \textbf{86.00} & 56.62 & 61.30 & 42.00 & 59.30 & \underline{78.22}\\\hline
EOGHorizontalSignal & 53.90 & \textbf{63.97} & 40.10 & 37.30 & 44.20 & \underline{57.73}\\\hline
EOGVerticalSignal & 50.30 & 47.76 & 37.60 & 29.80 & 39.20 & \textbf{51.10}\\\hline
Haptics & \textbf{52.60} & \underline{51.01} & 39.60 & 35.70 & 47.40 & 50.32\\\hline
Mallat & 91.40 & \underline{96.38} & 92.20 & 71.30 & 87.10 & \textbf{97.10}\\\hline
MixedShapesRegularTrain & 91.70 & \textbf{96.09} & 85.50 & 87.90 & 91.10 & \underline{93.69}\\\hline
MixedShapesSmallTrain & 86.10 & \textbf{91.79} & 73.50 & 82.80 & 81.30 & \underline{87.96}\\\hline
StarLightCurves & 96.90 & \underline{97.51} & 96.70 & 94.90 & 96.80 & \textbf{97.78}\\\hline\hline
Average & \underline{76.16} & 74.49 & 65.35 & 58.69 & 68.07 & \textbf{78.62}\\\hline
\end{tabular}
}
\label{tab:longTS}
\end{table}

\subsection{Quantitative Evaluation}\label{sec:quantitative}
We present the performance of the approach on the UCR-85 archive in Table~\ref{tab:sota}. We outperform other approaches on the benchmark while offering competitive performance to those designed to work in an ensemble. Interestingly, most state-of-the-art techniques are based on CNNs, with much effort spent finding optimal receptive field sizes for learning robust features at multiple timescales. Sequence-based approaches, such as those based on Transformers and RNNs, have struggled in this benchmark, mostly due to the limited training examples in many datasets. We, however, significantly outperform other sequence-based approaches and provide improvements of almost $5.5\%$ in absolute accuracy points over the closest RNN-based approach (TNC~\cite{tonekaboni2020unsupervised}). It also provides the best performance (out of non-ensemble approaches) on 17 datasets (also called \textit{wins} in prior literature~\cite{tang2021omni}) out of the 85 benchmark datasets. Additionally, it has an average rank of 5.35, performing competitively with other non-ensemble approaches. Ensemble models outperform all non-ensemble models by explicitly modeling sequential data by representing the sequential data at different time scales. However, they introduce additional overhead for handcrafting and fine-tuning multiple models.

\textbf{Performance on longer sequence data.} While the overall UCR-85 archive performance is excellent, we also examine the ability of the proposed approach to capture long-range dependencies when presented with time series data of longer durations. We select a subset of the UCR-128 archive, which contains additional datasets of longer duration. Specifically, we select 15 datasets with more than 1000 timesteps per sample without incomplete data. Table~\ref{tab:longTS} presents a summary of the results. As can be seen, we provide competitive performance with top-performing baselines with different backbone architectures. We have an average accuracy of 78.62\%, an average rank of 1.72, and provide ``wins'' in 6 out of the 15 long sequence datasets. It significantly improves over transformer-based (TS-TCC and TST) and RNN-based (TNC) baselines, which are trained to specifically model longer sequences through specialized training procedures such as contrastive learning. These results indicate the approach can capture robust representations from long sequences without complex ensemble processing. 

\textbf{Constrained Hardware Requirements} Our approach is designed to be simple and lightweight for use in settings with constrained training requirements, such as time and space budgets (i.e., limited training time, constrained hardware requirements, and limiting the number of parameters). Our model achieves competitive performance with 440k parameters and converges training on all datasets in 4 hours (on average over ten runs). For comparison, the current non-ensemble state-of-the-art approaches, TS2Vec (637k parameters) and ResNet (479k parameters), have more parameters and take longer to converge on a constrained hardware setup (32-core AMD ThreadRipper and NVIDIA RTX 3060). Similarly, on average, the BIC-based tokenization process (Section~\ref{eqn:BIC}) takes 500 ms for a sequence of 1000 data points, running in a single-threaded CPU-only application while having significantly less overhead for storing the components compared with other approaches. 

\begin{table}[t]
\centering
\caption{\textbf{Ablation studies} on the UCR-85 archive~\cite{bagnall2017great} to assess the impact of each component on the overall performance.
}
\begin{tabular}{|c|c|c|c|c|}
\hline
 \textbf{Backbone} & \textbf{MS-TSCS} & \textbf{$\mathcal{L}_{mae}$} & $\textbf{$\mathcal{L}_{CE}$}$ & \textbf{Accuracy}\\ \hline
Bi-LSTM & \ding{51} & \ding{51} & \ding{51} & \textbf{83.31}\\\hline
Bi-LSTM & \ding{55} & \ding{51} & \ding{51} & 73.68\\\hline
Bi-LSTM & \ding{51} & \ding{55} & \ding{51} & 75.31\\\hline
Bi-LSTM & \ding{51} & \ding{51} & \ding{55} & 74.28\\\hline
Bi-LSTM & \ding{55} & \ding{55} & \ding{51} & 68.33 \\\hline
Bi-RNN & \ding{51} & \ding{51} & \ding{51} & 81.54 \\\hline
Uni-LSTM & \ding{51} & \ding{51} & \ding{51} & 79.55\\\hline
\end{tabular}
\label{tab:ablation}
\end{table}

\textbf{Ablation Studies.} We systematically examine the impact of each module and summarize the results in Table~\ref{tab:ablation}. Specifically, we assess the effects of the multi-scale component discovery module (Section~\ref{sec:BIC}) and the choice of encoder model (Section~\ref{sec:BiLSTM}). Removing the component discovery model and using a fixed number of components for all datasets (set to 25, the median number of components across datasets) significantly hurts the performance. We also evaluate the strength of the learned representations by using a kNN instead of end-to-end training by removing $\mathcal{L}_{CE}$ from Equation~\ref{eqn:overall_loss}. 
While the loss in performance is expected, it does perform decently, indicating that the unsupervised loss function helps learn robust features. 
Removing $\mathcal{L}_{mae}$ results in significantly worse performance. 
Using bidirectional LSTMs instead of unidirectional LSTMs helps capture context and provides a more robust performance across all 85 datasets in the UCR archive. 
\begin{table}[t]
\centering
\caption{Evaluation of the BIC-based tokenization approach on the \textbf{time series segmentation} task~\cite{gharghabi2017matrix}.}
    \begin{tabular}{|c|c|c|c|}
    \hline
         \multirow{2}{*}{\textbf{Approach}} & \textbf{Learning} & \textbf{Pre-Defined}& \textbf{Mean}\\
          & \textbf{Phase?} & \textbf{Window?} & \textbf{Covering}\\\hline
         BinSeg & \ding{55} & \ding{51} & $52.4\pm 30.6$ \\ \hline
         PELT & \ding{55} & \ding{51} & $50.4 \pm 30.0$\\ \hline
         Window & \ding{55} & \ding{51} & $53.8 \pm 12.9$\\ \hline
         BOCD & \ding{55} & \ding{51} &  $55.5 \pm 14.4$\\\hline
         ESPRESSO & \ding{55} & \ding{51} & $58.0 \pm 15.8$ \\ \hline
         Ours & \ding{55} & \ding{55} & $\mathbf{72.7 \pm 12.5}$\\\hline\hline
         FLOSS & \ding{51} & \ding{51} &  $79.0 \pm 17.2$ \\\hline
         ClaSP & \ding{51} & \ding{55} &  $\mathbf{79.8 \pm 20.4}$ \\\hline\hline
         Ours & \ding{55} & \ding{51} & $\mathbf{78.3 \pm 12.9}$\\\hline
    \end{tabular}
    \label{tab:segmentation_utsa}
\end{table}

\section{Extension to Unsupervised Time Series Segmentation}\label{sec:segmentation}

In addition to evaluating the performance of our approach on time series classification, we assess the quality of the components obtained through the BIC-based segmentation (Section~\ref{sec:BIC}) by evaluating it on the time series segmentation task~\cite{gharghabi2017matrix}. The goal of time series segmentation is to identify natural segments caused by \textit{change points} in sequential data where there are sudden changes in statistical properties of the time series due to changes in events captured by the data. For example, these changes could point to transitions between actions performed by a subject. The UTSA benchmark~\cite{gharghabi2017matrix} introduces a set of 32 datasets derived from the UCR archive~\cite{bagnall2017great} and provides human-annotated segments of datasets across 16 different use cases from biological, mechanical, and synthetic processes. Each use case in the benchmark contains, on average, 2 to 3 segments derived from real, semi-synthetic, and artificial changes and provides a considerable challenge for unsupervised time series segmentation. 
 
We use the components discovered using the multi-scale change space model as segments and assess the quality of the segmentations on the UTSA benchmark. We compare against a variety of baselines such as BinSeg~\cite{sen1975tests}, PELT~\cite{killick2012optimal}, Window~\cite{truong2020selective}, BOCD~\cite{adams2007bayesian}, ESPRESSO~\cite{deldari2020espresso}, FLOSS~\cite{gharghabi2019domain}, and ClaSP~\cite{schafer2021clasp}, which represent the commonly used state-of-the-art unsupervised segmentation approaches. We use the mean covering with standard deviation as a metric to quantify the performance of the approaches. Based on the Jaccard index, the covering score provides a weighted overlap between the ground truth and the predicted segments. Higher values indicate better alignment between the predicted and the ground truth segments. We report results from the implementations from ClaSP~\cite{schafer2021clasp} for a fair comparison and consistent experimental setup.

\begin{figure*}
    \begin{tabular}{cc}
    \hline
    \multicolumn{2}{c}{\textbf{Segmentation Examples}}
    \\\hline
     \includegraphics[width=0.45\columnwidth]{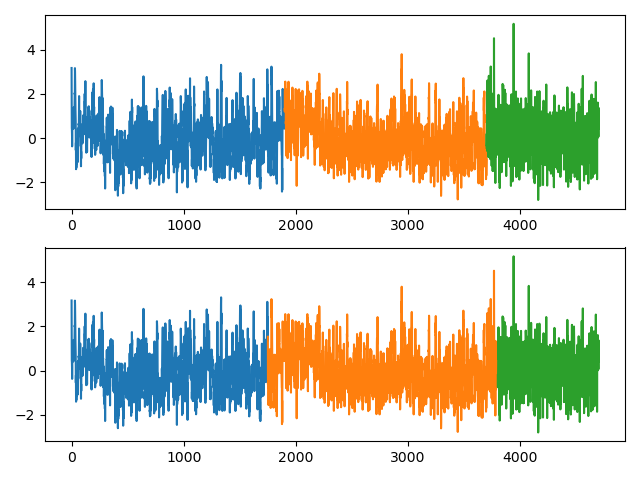}    &  \includegraphics[width=0.45\columnwidth]{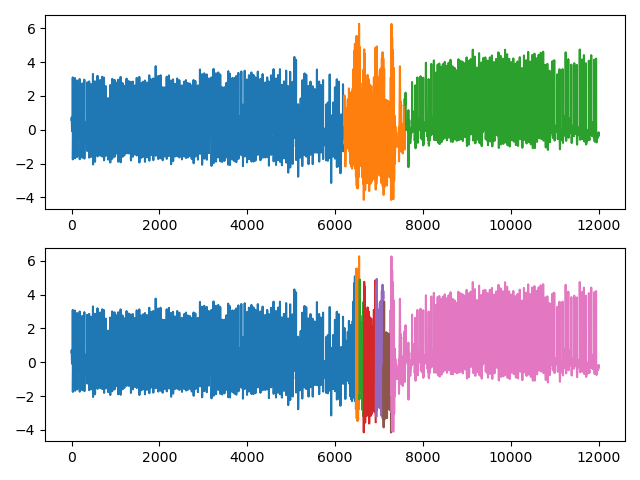} \\
     (a) & (b)\\
    \end{tabular}
    \caption{\textbf{Qualitative visalization} of (a) a successful segmentation and (b) unsuccessful segmentation on the GreatBarbet2 and SuddenCardianDeath1 datasets, respectively. The first row shows ground truth segments, and the second shows predicted segments.}
    \label{fig:viz}
\end{figure*}

Table~\ref{tab:segmentation_utsa} summarizes the results. We significantly outperform other non-learning-based approaches that require a pre-defined period size (temporal window) corresponding to the ideal time scale at which the change points can be detected reliably. 
This value is often domain-dependent and requires extensive handcrafting (of architecture or features) to capture, especially in time series classification and segmentation. Our approach can automatically search for this using the multi-scale change space and considers change points at different temporal granularities. 
When given this optimal window, we establish the change space at this time scale and perform segmentation. 
As can be seen, we perform competitively with learning-based approaches and further widen the gap with the non-learning-based approaches. Interestingly, we perform exceptionally well without the optimal time scale, indicating that the multi-scale change space captures the change points at time scales approaching the ideal scale. Some example segmentations are shown in Figure~\ref{fig:viz}, where it can be seen that our approach can segment signals into their components without training and supervision. Although it over segments in some instances, the segments are statistically meaningful, are captured at multiple time scales, and do not always correspond to the ground change points extracted at a single time scale. 
For example, in Figure~\ref{fig:viz}(b), we see that over-segmentation occurs during periods of intense changes and captures fine-grained change points but has excellent coverage during stable regions on either side of this rapidly changing segment. 
Note that our approach detects the temporal components in a time-scale and class-agnostic manner and does not have access to the ideal time scale at which the ground truth is annotated. Despite this over-segmentation, it allows us to capture robust features for classification. 

\section{Discussion and Future Work}\label{sec:conclusion}
In this work, we presented a novel multi-scale change-space approach to discover temporal components in univariate time series data and provide an intuitive way to tokenize time series data using statistical measures. Given these components, we learn compositional representations using sequence-based encoders by training the model as a masked, denoising auto-encoder. Evaluation on 85 publicly available datasets on the benchmark UCR-85 archive demonstrates its effectiveness in learning robust representations. Additional experiments on segmentation benchmarks demonstrate that the detected components are highly correlated with naturally occurring segments found in time series data. We aim to extend this formulation to capture part-whole hierarchies for learning \textit{hierarchical} compositional representations from multi-modal and multi-variate time series data with longer temporal durations. 

\textbf{Acknowledgements.} 
This work was supported by the U.S. National Science Foundation Grant IIS 2348689 and IIS 2348690 and U.S. Department of Agriculture Grant 2023-69014-39716-1030191. 
\bibliographystyle{splncs04}
\bibliography{refs}
\end{document}